\documentclass[runningheads]{llncs}
\usepackage[T1]{fontenc}

\usepackage{graphicx}

\usepackage{cite}
\usepackage{amsmath,amssymb,amsfonts}
\usepackage{algorithmic}
\usepackage{graphicx}
\usepackage{textcomp}
\usepackage{xcolor}

\usepackage{doc}
\usepackage{algorithm}
\usepackage{xspace}
\usepackage{graphicx,subfigure}
\usepackage{caption}
\usepackage{paralist}
\usepackage{amsfonts} 
\usepackage{xcolor}
\usepackage{subfigure}
\usepackage{wrapfig}
\usepackage{url}
\newcommand{\mysubsection}[1]{\medskip\noindent\textbf{#1}}
\newcommand{\ourTool}{gRoMA}

\DeclareMathOperator*{\argmmax}{arg\,max}

\begin{document}
\title{\ourTool{}: a Tool for Measuring the Global Robustness of Deep Neural Networks}
\author{Natan Levy\inst{1} \and Raz Yerushalmi\inst{1, 2}\orcidID{0000-0002-0513-3211} \and Guy Katz\inst{1}}

\authorrunning{}

\institute{The Hebrew University of Jerusalem, Jerusalem, Israel
	\email{\{natan.levy1,gkatz\}@mail.huji.ac.il}\\
	 \and The Weizmann Institute of Science, Rehovot, Israel \email{raz.yerushalmi@weizmann.ac.il}}
\maketitle              
\begin{abstract}
Deep neural networks (DNNs) are at the forefront of cutting-edge
technology, and have been achieving remarkable performance in a
variety of complex tasks. Nevertheless, their integration into
safety-critical systems, such as in the aerospace or automotive
domains, poses a significant challenge due to the threat of
\emph{adversarial inputs}: perturbations in inputs that might cause
the DNN to make grievous mistakes. Multiple studies have
demonstrated that even modern DNNs are susceptible to adversarial
inputs, and this risk must thus be measured and mitigated to allow
the deployment of DNNs in critical settings.
Here, we present \ourTool{} (global Robustness Measurement and
Assessment), an innovative and scalable tool that implements a probabilistic 
approach to measure the global
categorial robustness of a DNN. Specifically, \ourTool{}
measures the probability of encountering adversarial inputs for a
specific output category. Our tool operates on pre-trained,
black-box classification DNNs, and generates input samples belonging
to an output category of interest. It measures the DNN's
susceptibility to adversarial inputs around these inputs, and
aggregates the results to infer the overall global categorial
robustness of the DNN up to some small bounded statistical
error.
We evaluate our tool on the popular Densenet DNN model over the CIFAR10 dataset. Our results reveal significant gaps in the robustness of the different output categories. This experiment demonstrates the usefulness and scalability of our approach and its potential for allowing DNNs to be deployed within critical systems of interest.

\end{abstract}
\section{Introduction}

Deep neural networks (DNNs) have become fundamental components in many
applications that perform classification~\cite{KrHi09, AlTaTa17}.
Empirically, DNNs often outperform traditional software, and even
humans~\cite{PaDaJo03, XuChHeZhDu19}. Nevertheless, DNNs have a
significant drawback: they are notoriously susceptible to small input
perturbations, called \emph{adversarial inputs}~\cite{GoShSz15}, which
can cause them to produce erroneous outputs. These adversarial inputs
are one of the causes likely to delay the adoption of DNNs in
safety-critical domains, such as aerospace~\cite{GaShTiWi23},
autonomous vehicles~\cite{LaHuWaLiSuZh18}, and medical
devices~\cite{HaLeWi22}.

In the aforementioned critical domains, systems must meet high
dependability standards. While strict guidelines exist for certifying
that hand-crafted software meets these standards (e.g., the DO-178
standard~\cite{FAA93} in the aerospace industry), no such
certification guidelines currently exist for systems incorporating
DNNs. Several regulatory agencies have recognized the existence of
this gap and the importance of addressing it. For example,
in its recently published roadmap,
the European Union Aviation Safety Administration (EASA) has
emphasized the importance of DNN robustness as one of the 7 key
requirements for trustworthy artificial intelligence~\cite{EuUnAvSaAg23}. However, certifying the robustness of DNNs remains an open problem.

The formal methods community has begun addressing this gap by
devising methods for rigorously quantifying the \emph{local
	robustness} of a DNN~\cite{KaBaDiJuKo17, WaPeWhYaJa18,
	ZhZhChSoCh21}. Local robustness refers to a DNN's ability to
withstand adversarial inputs in the vicinity of a specific point
within the input space. Although the rigorous verification
approaches proposed to date have had some success in measuring these 
robustness scores, they typically struggle to scale as network sizes
increase~\cite{KaBaDiJuKo17} --- which limits their practical
application. To circumvent that limitation, \emph{approximate} methods
have been proposed, which can evaluate DNN robustness more
efficiently, but often at the cost of reduced
precision~\cite{TiFuRo21, MaNoOr19, HuHuHuPe21, CoRoZi19, BaChMeSa21}.

Work to date, both on rigorous and on approximate methods, has focused
almost exclusively on measuring \emph{local} robustness, which
quantifies the DNN's robustness around individual input points within
a multi-dimensional, infinite input space. In the context of DNN
certification, however, a broader perspective is required --- one that
measures the \emph{global robustness} of the DNN, over the entire
input space, rather than on specific points.

In this paper, we propose a novel approach for approximating the
global robustness of a DNN. Our method is computationally
efficient, scalable, and can handle various types of adversarial
attacks and black-box DNNs. Unlike existing approximate approaches,
our approach provides statistical guarantees about the precision of the
computed robustness score.

More concretely, our approach (implemented in the \ourTool{} tool)
implements a probabilistic verification approach for performing global
robustness measurement and assessment on DNNs. \ourTool{} achieves
this by measuring the \emph{probabilistic global categorial robustness} (\emph{PGCR}) of a given DNN. In this study, we take a
conservative approach and consider the DNN as a black-box: \ourTool{}
makes no assumptions, e.g., about the Lipschitz continuity of the DNN,
the kinds of activation functions, the hyperparameters it uses, or its
internal topology. Instead, \ourTool{} uses and extends the recently
proposed RoMA (\emph{a Method for DNN Robustness Measurement and
	Assessment}) algorithm~\cite{LeKa21} for measuring local
robustness. \ourTool{} repeatedly invokes this algorithm on a
collection of samples, drawn to represent a specific output category
of interest; and then aggregates the results to compute a global
robustness score for this category, across the entire input space. As
a result, \ourTool{} is highly scalable, typically taking only a few
minutes to run, even for large networks. Further, the tool formally
computes an error bound for the estimated PGCR scores using
Hoeffding's inequality \cite{Ho63} to mitigate the drawbacks of using
a statistical method. Thus, \ourTool{}'s results can be used in the
certification process for components of safety-critical systems,
following, e.g., the SAE Aerospace Recommended Practice~\cite{LaNi11}.

For evaluation purposes, we focused on a Densenet DNN~\cite{HuLiVaWe17}, trained on the CIFAR10 dataset~\cite{KrHi09}; and
then measured the network's global robustness using one hundred
arbitrary images for each CIFAR10 category. \ourTool{} successfully
computed the global robustness scores for these categories,
demonstrating, e.g., that the airplane category is significantly more
robust than other categories.

To the best of our knowledge, our tool is presently the only scalable
solution for accurately measuring the \emph{global categorial
  robustness} of a DNN, i.e., the aggregated robustness of \emph{all}
points within the input space that belong to a category of interest
--- subject to the availability of a domain expert who can supply
representative samples from each category. The availability of such
tools could greatly assist regulatory authorities in assessing the
suitability of DNNs for integration into safety-critical systems, and
in comparing the performance of multiple candidate DNNs.

\mysubsection{Outline.} We begin with an overview of related work on
measuring the local and global adversarial robustness of DNNs, in Section~\ref{RelatedWork}. In Section~\ref{Background}, we provide the necessary definitions for understanding our approach. We then introduce the~\ourTool{} tool in Section~\ref{Introducing}. Next, in Section~\ref{Evaluation}, we evaluate the performance of our tool using a popular dataset and DNN model. Finally, Section~\ref{Conclusion} concludes our work and discusses future research directions.

\section{Related Work}
\label{RelatedWork}
Measuring the local adversarial robustness of DNNs has received
significant attention in recent years. Two notable approaches for addressing it are:
\begin{itemize}
	\item Formal-verification approaches~\cite{MuMaSiPuVe22,
		WaZhXuLiJaHsKo21, KaHuIbJuLaLiShThWuZeDiKoBa19}, which utilize
	constraint solving and abstract interpretation techniques to
	determine a DNN's robustness. These approaches are fairly precise,
	but generally afford limited scalability, and are applicable only to
	white-box DNNs.
	
	\item Statistical approaches, which evaluate the probability of
	encountering adversarial inputs. These approaches often need to
	balance between scalability and accuracy, with prior work~\cite{TiFuRo21,
		MaNoOr19, HuHuHuPe21, CoRoZi19, BaChMeSa21} typically leaning
	towards scalability.
\end{itemize}

Recently, the \emph{RoMA} algorithm~\cite{LeKa21} has been introduced
as highly scalability  statistical method, but which can also provide
rigorous guarantees on accuracy. RoMA is a simple-to-implement algorithm
that evaluates local robustness by sampling around an input point of
interest; measuring the confidence scores assigned by the DNN to
\emph{incorrect} labels on each of the sampled input points; and then using
this information to compute the probability of encountering an input
on which the confidence score for the incorrect category will be high
enough to result in misclassification. In the final step, RoMA
assesses robustness using properties of the normal distribution
function~\cite{LeKa21}. RoMA handles black-box DNNs, without any a
priori assumptions; but it can only measure local, as opposed to
global robustness.

Due to the limited usefulness of computing local robustness in modern
DNNs, initial attempts have been made to compute the \emph{global
	adversarial robustness} of networks. Prior work formulated and
defined the concept of \emph{global adversarial
	robustness}~\cite{KaBaDiJuKo17, MaNoOr19}; but in the same breath,
noted that global robustness can be hard to check or compute compared
to local robustness. More recently, there have been attempts to use
formal verification to check global adversarial
robustness ~\cite{WaWaFuJiHuLiZh22, ZhWeZhXiMe23,KhSh23}; but the reliance on
formal verification makes it difficult for these approaches to scale,
and requires a white-box DNN with specific activation functions.

Two other recently proposed approaches study an altered version of
global robustness. The first work, by Ruan et
al.~\cite{RuWuSuHuKrKw19}, defines global robustness as the expected
maximal safe radius around a test data set. It then proposes an
approximate method for computing lower and upper bounds on DNN's
robustness. The second work, by Zaitang et al.~\cite{ZaChHo23},
redefines global robustness based on the probability density function,
and uses generative models to assess it. These modified definitions
of global robustness 
present an intriguing perspective. However, it is important to note that
they differ from common definitions, and whether they will be widely adopted remains to be seen.
Another noteworthy recent approach, proposed by Leino et
al.~\cite{LeWaFR21}, advocates for training DNNs that are certifiably
robust by construction, assuming that the network is Lipschitz-continuous. 
This approach can guarantee the global robustness of a DNN without accurate measurements, but it requires the DNN to be white-box, whereas our approach is also compatible with black-box DNNs.

Our work here focuses on measuring and scoring the global robustness of pre-trained black-box DNNs and is the first, to the best of our knowledge, that is scalable and consistent with the commonly accepted definitions.

\section{DNNs and Adversarial Robustness}
\label{Background}

\mysubsection{Neural Networks.} A DNN $N: \mathbb{R}^n \rightarrow \mathbb{R}^m$ is a function that maps input $\vec{x} \in \mathbb{R}^n$ to output $\vec{y} \in \mathbb{R}^m$. In classifier DNNs, which are our subject matter here, $\vec{y}$ is interpreted as a vector of confidence scores, one for each of $m$ possible labels. We say that $N$ classifies $\vec{x}$ as label $l$ iff $\argmmax(\vec{y})=l$, i.e., when $y$'s $l$'th entry has the highest score. We use $L$ to denote the set of all possible labels, $L=\{1,\ldots,m\}$.

\mysubsection{Local Adversarial Robustness.} The local adversarial
robustness of $N$ around input $\vec{x}$ is a measure of how
sensitive $N$ is to small perturbations around $\vec{x}$~\cite{BaIoLaVyNoCr16}:
\begin{definition}
	\label{definition1}
	A DNN $N$ is $\epsilon$-locally-robust at input point $\vec{x_0}$ iff
	\[
	\forall \vec{x}.\quad
	\displaystyle || \vec{x} -\vec{x_0} ||_{\infty} \le \epsilon 
	\Rightarrow \argmmax(N(\vec{x})) = \argmmax(N(\vec{x_0})) 
	\]
\end{definition}
Intuitively, Definition~\ref{definition1} states that the network
assigns to $\vec{x}$ the same label that it assigns to $\vec{x_0}$,
for input $\vec{x}$ that is within an $\epsilon$-ball around
$\vec{x_0}$. Larger values of $\epsilon$ imply a larger ball around
$\vec{x_0}$, and consequently --- higher robustness.

The main drawback in Definition~\ref{definition1} is that it considers
a single input point in potentially vast input space. Thus, the
 $\epsilon$-local-robustness of $N$ at $\vec{x_0}$ does not imply that
it is also robust around other points. Moreover, it assumes that DNN robustness is consistent across categories, although it has already been observed that some categories can be more robust than others~\cite{LeKa21}. To overcome these drawbacks, the notion of \emph{global categorial robustness} has been proposed~\cite{KaBaDiJuKo17b, RuWuSuHuKrKw19}:

\begin{definition}
	\label{definition2} 
	A DNN $N$ is ($\epsilon,\delta$)-globally-robust in input region $D$ iff
	\[
	\forall \vec{x_1},\vec{x_2} \in D.
	\]
	\[ 
	\displaystyle || \vec{x_1} -\vec{x_2} ||_{\infty} \le \epsilon 
	\Rightarrow  
	\forall l \in L.~ | N(\vec{x_1})[l] - N(\vec{x_2})[l] | < \delta
	\]
\end{definition}

Intuitively,
Definition~\ref{definition2} states that for every two inputs
$\vec{x_1}$ and $\vec{x_2}$ that are at most $\epsilon$ apart, there are no spikes greater than $\delta$ in the confidence scores that the DNN assigns to each of the labels.

Definitions~\ref{definition1} and~\ref{definition2} are Boolean in
nature: given $\epsilon$ and $\delta$, the DNN is either robust or not
robust. However, in real-world settings, safety-critical systems can
still be determined to be sufficiently robust if the \emph{likelihood}
of encountering adversarial inputs is sufficiently
low~\cite{LaNi11}. Moreover, it is sometimes more appropriate to
measure robustness for specific output categories~\cite{LeKa21}.  To
address this, we propose to compute real-valued, \emph{probabilistic
  global categorial robustness} scores:

\begin{definition}
	\label{definition3}
	Let $N$ be a DNN, let $l\in L$ be an output label, and let $I$ be a finite set of labeled data representing the input space for $N$. 
	The ($\epsilon,\delta$)-PGCR score for $N$ with respect to $l$ and
	$I$, denoted
	$pgcr_{\delta,\epsilon}(N,l,I)$,
	is defined as:
	\[
	pgcr_{\delta,\epsilon}(N,l,I) \triangleq  
	P_{\vec{x_1}\in I,\vec{x_2}\in \mathbb{R}^n || \vec{x_1} -\vec{x_2} ||_{\infty} \le \epsilon}
	[| N(\vec{x_1})[l] - N(\vec{x_2})[l] | < \delta ] 
	\]	
\end{definition}

Intuitively, the definition captures the probability that for an
input $\vec{x_1}$ drawn from $I$, and for an additional input
$\vec{x_2}$ that it is at most $\epsilon$ apart from $\vec{x_1}$, inputs $\vec{x_1}$ and $\vec{x_2}$ will be assigned confidence scores that differ by at most $\delta$
for the label $l$.

\section{Introducing the \ourTool{} Tool}
\label{Introducing}

\mysubsection{Algorithm.} The high-level flow of \ourTool{} implements
Definition~\ref{definition3} in a straightforward and efficient way: it first computes the local robustness for $n$ representative points from each category, and then aggregates the global robustness using Algorithm~\ref{algorithm1}.

The inputs to \ourTool{} are:
\begin{inparaenum}[(i)]
	\item a network $N$; 
	\item $I$, a finite set of labeled data that represents the input space, to draw samples from;
	\item a label $l$;
	\item $n$, the number of representative samples of inputs classified
	as $l$ to use; and
	\item $\epsilon$ and $\delta$, which determine the allowed
	perturbation sizes and differences in confidence scores, as per Definition~\ref{definition3}. 
\end{inparaenum}
\ourTool{}'s output consists of the computed $pgcr_{\delta,\epsilon}(N,l,I)$ score and an error term $e$, both specific to $l$. 
We emphasize the reliance of the $pgcr_{\delta,\epsilon}$ score on having representative input samples for each relevant category $l$.
Under that assumption, in which the samples represent the underlying input distribution, our method guarantees that, with some high, predefined probability, the distance of the computed $pgcr_{\delta,\epsilon}$ value from its true value is at most $e$.

\begin{algorithm}[ht]
	\caption{\ourTool($N , I, l , n ,
		\epsilon, \delta$)}
	\label{algorithm1}
	\begin{algorithmic}[1]
		\STATE
		$\vec{X}$ := drawSamples($I,l,n$) \label{line:pickSamples}
		\FOR {$i:=1$ to $n$} 
		\IF { ( $N(\vec{X}[i])=l$ )}\label{line:CheckClassification}
		\STATE $\vec{plr[i]}$ :=
		RoMA($\vec{X}[i],\epsilon,\delta,N$) \label{line:applyRoma}
		\ENDIF
		\ENDFOR
		\STATE pgcr := aggregate($\vec{plr}$) \label{line:aggregate}
		\STATE $e$ :=
		computeError(pgcr, $\vec{plr}$, $\vec{X}$) \label{line:errorBound}
		\RETURN (pgcr,$e$) \label{line:return}
		
	\end{algorithmic}
\end{algorithm}

In line~\ref{line:pickSamples}, \ourTool{} begins by creating a
vector, $\vec{X}$, of perturbed inputs --- by drawing from $I$, at
random, $n$ samples of inputs labeled as $l$. Next, for each 
correctly classified sample (line~\ref{line:CheckClassification}), \ourTool{} computes the sample's
probabilistic local robustness (\emph{plr}) score using
RoMA~\cite{LeKa21} (line~\ref{line:applyRoma}). Finally, \ourTool{}
applies statistical aggregation (line~\ref{line:aggregate}) to compute
the $pgcr$ score and the error bound (line~\ref{line:errorBound});
and these two values are then returned on line~\ref{line:return}.

\ourTool{} is modular in the sense that any aggregation method
(line~\ref{line:aggregate}) and error computation method
(line~\ref{line:errorBound}) can be used. 
There are several suitable techniques in the statistics literature for
both tasks, a thorough discussion of which is beyond our scope
here. We focus here on a few straightforward mechanisms for these tasks, which we describe next.

For score aggregation, we propose to use the numerical average of the
local robustness scores computed for the individual input samples.
Additional approaches include computing a median score and more
complex methods, e.g., methods based on normal distribution
properties \cite{Ca01}, maximum likelihood methods, Bayesian
computations, and others. For computing the PGCR score's probabilistic error bound,
we propose to use Hoeffding's Inequality \cite{Ho63}, which provides an upper
bound on the likelihood that a predicted value will
deviate from its expected value by more than some specified amount.

\section{Evaluation} 
\label{Evaluation}
\mysubsection{Implementation.} We implemented \ourTool{} as a Tensorflow framework~\cite{Tensorflow2}. Internally, it uses Google Colab~\cite{Colab, BiBi19} tools with 12.7GB system RAM memory, and T4 GPU. It accepts DNNs in Keras H5 format~\cite{Keras15}, as its input.
The \ourTool{} tool is relatively simple, and can be extended
and customized to support, e.g., multiple input distributions of
interest, various methods for computing aggregated robustness scores
and probabilistic error bounds, and also to accept additional DNN
input formats. \ourTool{} is available online~\cite{gRoMA}.

\mysubsection{Setup and Configuration.} 
We conducted an evaluation of \ourTool{} on a commonly used Densenet
model~\cite{HuLiVaWe17} with 797,788 parameters, trained on the
CIFAR10 dataset~\cite{KrHi09}. The model achieved a test accuracy of
93.7\% after a standard 200-epoch training period. The code for
creating and training the model, as well as the H5 model file, are available online~\cite{gRoMA}

For \ourTool{} to operate properly, it is required to obtain a representative
sample of the relevant input space $I$. Creating such a representative
sample typically requires some domain-expert knowledge~\cite{Om14,
  GrSc14}. However, random sampling can often serve as an
approximation for such sampling~\cite{Om14, GrSc14}; a more thorough discussion of
that topic goes beyond the scope of this paper. In our experiments
here, we used a simple sampling mechanism in order to demonstrate the
use of \ourTool{}.
We measured the global categorial robustness of each output category
by running 
the \emph{RoMA} algorithm~\cite{LeKa21}, to calculate the 
local robustness of one hundred images drawn independently and arbitrarily from the set $I$, which includes varying angles, lighting conditions, and resolutions. We set
$\epsilon$ to 0.04 and $\delta$ to 0.07, as recommended in that 
work~\cite{LeKa21}. 

Due to our desire to check the approach's
applicability to the aerospace industry, we paid special attention to
the airplane category. In this category, we focused on Airbus A320-200
commercial airplane images, either airborne or on the ground. This
type of airplane exists in the CIFAR10 training set as well, and hence we expected a high level of categorial global robustness for this category.
The images, along with our code and dependencies, are available online~\cite{gRoMA}.

Next, for each output category, we used \emph{RoMA} to compute the \emph{probabilistic local robustness (plr)} score for each input sample. We configured \ourTool{} to use the numerical average as the aggregation method; and for assessing the error of \ourTool{}, we applied Hoeffding's inequality~\cite{Ho63}.
Specifically, we aimed for a maximum expected error value of 5\%,
which is an acceptable error value when calculating a DNN's
robustness~\cite{HuHuHuPe21}. We used Hoeffding's inequality to
calculate the probability that the actual error is higher than this
value. This was achieved by setting the upper and lower bounds of the
$plr$ values to be plus and minus five standard deviations of the
$plr$ values, corresponding to a $1-1*10^{-6}$ accuracy, all in a
normal distribution context.
These bounds were selected in order to 
 provide a conservative estimate, encompassing a significant portion
 of the input space.
We justify the normal distribution assumption using Anderson-Darling goodness-of-fit test~\cite{An11} that focuses on the tails of the distribution~\cite{BeKoSc21}, as detailed in \cite{LeKa21}.

\mysubsection{Results.} Running our evaluation took less than 21
minutes for each category, using a Google Colab~\cite{Colab} machine.
The various global robustness scores for each category, as well as the
calculated probabilistic error, appear in Fig.~1.

In the evaluation, the Airplane category obtained, as expected while focusing on a specific type of airplane, the
highest categorial robustness score of 99.91\% among all categories;
while the Cat and Ship categories obtained the lowest score of 99.52\%
(the PGCR scores appear in blue in Fig.~1). 
The statistical error margin (tolerance) was set to 5\% for this study. Based on these setting, the Ship category had the highest probability to exceed this bound, at less than 0.16\%. On the other hand, the Airplane, Automotive, Bird, Dog, and Frog categories had the lowest error likelihood, all below 0.0005\% (the likelihood scores per category scores appear in yellow in Fig.~1).

\begin{figure}[ht]
	\label{figure1categories}
	\begin{center}
		\centering
		\includegraphics[trim = 1.5cm 10cm 1cm 11.5cm,clip,width=.97\textwidth]{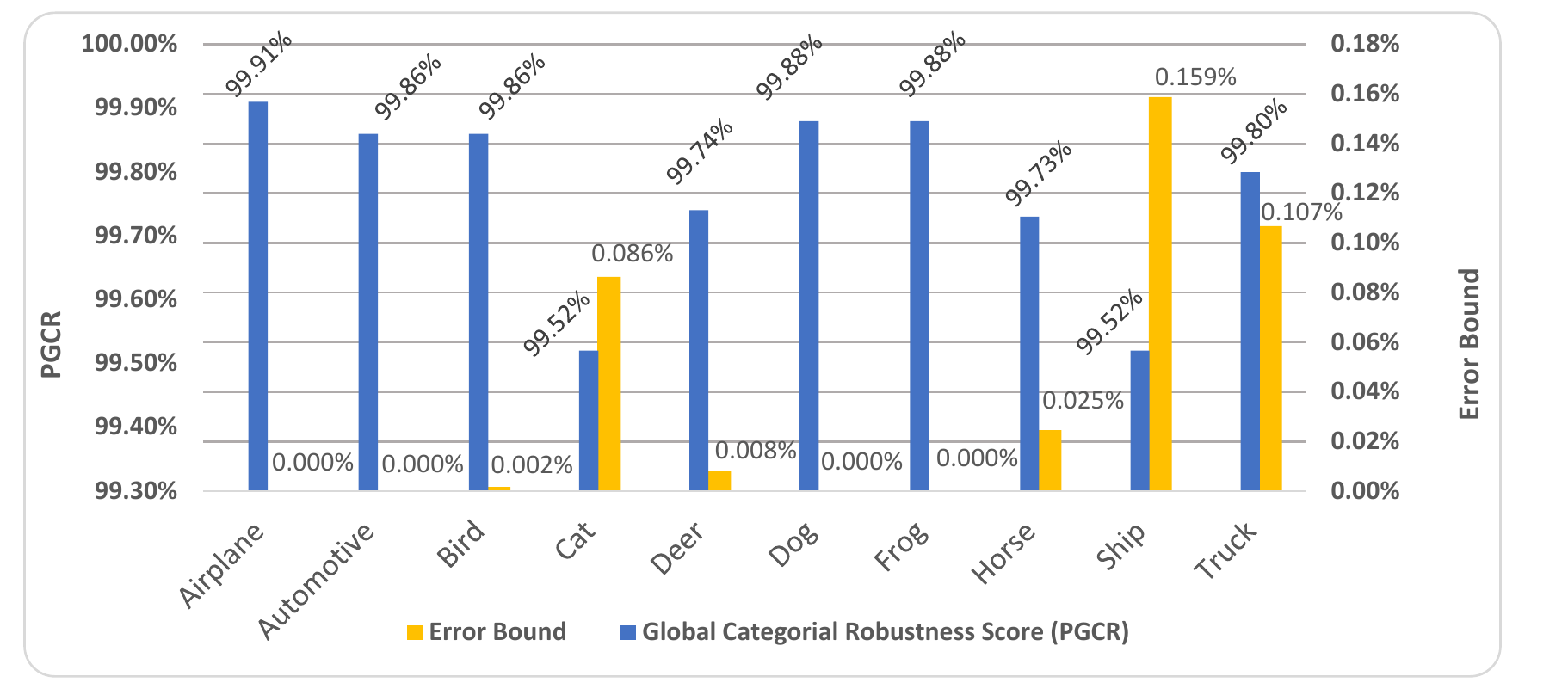} 
		\caption{PGCR scores, per category, for all CIFAR10
			categories (blue); and the corresponding statistical errors (yellow).}
	\end{center}	
\end{figure}

The PGCR scores calculated are aligned with previous research, that already assessed the local robustness of all the images in the CIFAR10 test set, and which indicate that different categories may obtain different robustness scores~\cite{LeKa21}. 

\section{Conclusion and Future Work}
\label{Conclusion}
We introduced here the notion of PGCR and presented the \ourTool{} tool for probabilistically measuring the global categorial robustness of DNNs, e.g., calculating the $pgcr_{\epsilon, \delta}$ score --- which is a step towards formalizing DNN safety and reliability for use in safety-critical applications. Furthermore, we calculate a bound on the statistical error inherent to using a statistical tool. The main contribution of this work is developing a scalable tool for probabilistically measuring categorial global DNN robustness.

Although extensive research has focused on DNN local adversarial
robustness, we are not aware of any other scalable tool that can
measure the global categorial robustness of a DNN. Therefore, we
believe that our tool provides a valuable contribution to the research community.

In future work, we plan to test the accuracy of \ourTool{} using a range of input distributions and sampling methods, simulating various input spaces used in different applications. Additionally, we intend to extend our tool to other types of DNNs, such as regression networks, to broaden PGCR's applicability. These efforts will enhance our understanding of DNN robustness and facilitate safe and reliable deployment in real-world applications.

\subsection{Acknowledgments.} We thank Dr.~Or Zuk of the Hebrew
University for his valuable contribution and support.  This work 
was partially supported by the Israel
Science Foundation (grant number 683/18). The work of Yerushalmi was
partially supported by a research grant from the Estate of Harry
Levine, the Estate of Avraham Rothstein, Brenda Gruss and Daniel
Hirsch, the One8 Foundation, Rina Mayer, Maurice Levy, and the Estate
of Bernice Bernath, grant 3698/21 from the ISF-NSFC, and a grant from the Minerva foundation.

 \bibliographystyle{splncs04}
 \bibliography{gRoMA}

\end{document}